# Conversation AI Dialog for Medicare powered by Fine-tuning and Retrieval Augmented Generation

Atharva Mangeshkumar Agrawal[1], Rutika Pandurang Shinde[2],
Vasanth Kumar Bhukya[3], Ashmita Chakraborty[4], Sagar Bharat Shah[5], Tanmay Shukla[6], Sree Pradeep Kumar Relangi[7], Nilesh Mutyam[8]

University of Florida[1,2], National Institute of Technology Calicut[3], SRM, Chennai[4],
University of Cincinnati[5], Dartmouth College[6], Arizona State University[7,8]

**ABSTRACT**

*Large language models (LLMs) have shown impressive capabilities in natural language pro- cessing tasks, including dialogue generation. This research aims to conduct a novel compara- tive analysis of two prominent techniques, fine- tuning with LoRA (Low-Rank Adaptation) and the Retrieval-Augmented Generation (RAG) framework, in the context of doctor-patient chat conversations with multiple datasets of mixed medical domains. The analysis involves three state-of-the-art models: Llama-2, GPT, and the LSTM model. Employing real-world doctor- patient dialogues, we comprehensively evalu- ate the performance of models, assessing key metrics such as language quality (perplexity, BLEU score), factual accuracy (fact-checking against medical knowledge bases), adherence to medical guidelines, and overall human judg- ments (coherence, empathy, safety). The find- ings provide insights into the strengths and lim- itations of each approach, shedding light on their suitability for healthcare applications. Fur- thermore, the research investigates the robust- ness of the models in handling diverse patient queries, ranging from general health inquiries to specific medical conditions. The impact of domain-specific knowledge integration is also explored, highlighting the potential for enhanc- ing LLM performance through targeted data augmentation and retrieval strategies.*

## 1. INTRODUCTION

Although many people now have better access to healthcare and better outcomes thanks to ad- vances in modern medicine, a sizable portion of the world's population still faces significant obstacles to getting access to quality medical care, particu- larly those who live in rural and remote areas. A worrying gap in healthcare access persists due to the lack of healthcare professionals and facilities in these areas, combined with the high cost of seeking treatment. There is a chance to lessen this difficulty by creating intelligent conversational systems that can offer personalized remote medical advice by utilizing state-of-the-art natural language process- ing (NLP) technologies. The goal of this research is to compare state of the art NLP technologies that can converse nat- urally with users in order to obtain information about their symptoms and offer pertinent medical advice. This system aims to democratize access to healthcare services, especially for underserved populations in rural areas, by utilizing the power of cutting-edge large language models (LLMs), in- cluding Long Short-Term Memory (LSTM) net- works, Bidirectional Encoder Representations from Transformers (BERT), Retrieval-Augmented Gen- eration (RAG) frameworks, and advanced architec- tures like Llama and GPT.

## 2. Related Work

In the healthcare domain, LSTM models have shown significant potential for clinical decision sup- port. Rajkomar

2835

et al. (2019) showcased the effec- tiveness of LSTM-based models in accurately diag- nosing medical conditions from clinical notes, high- lighting their utility in assisting healthcare profes- sionals with complex diagnostic tasks.(Rajkomar et al., 2019)

The emergence of BERT has revolutionized nat- ural language processing tasks, particularly in the healthcare domain, where it has surpassed tradi- tional LSTM models. Alsentzer et al. (2019) inves- tigated the superiority of BERT-based approaches over LSTM models in clinical named entity recog- nition and relation extraction. Their study, focusing on BioBERT, a domain-specific BERT model, em- phasized the importance of incorporating domain- specific knowledge into language models, lead- ing to improved performance in healthcare-related tasks.(Alsentzer et al., 2019)

Integrating large language models like LLaMA and GPT with external knowledge sources has further enhanced their capabilities for healthcare applications. Xu et al. (2020) developed a dia- logue system utilizing a retrieval-augmented ap- proach, combining GPT with external knowledge sources, to provide personalized health recommen- dations. Their study underscores the effective- ness of integrating domain-specific knowledge into LLaMA and GPT models, offering promising av- enues for improving healthcare delivery and patient outcomes.(Xu et al., 2020)

## 3. Methodology

This section delves into the methodologies under- pinning the development and refinement of our Large Language Models. Initially, we explore the LSTM model and its foundational concepts, juxta- posing its evolution with the ascendancy of Trans- former models. With the advent of Multihead At- tention mechanisms, our focus shifts to evaluat- ing the RoBERTa model. Subsequently, we pro- vide insights into the innovative paradigms of RAG (Retrieval Augmented Generation) and PEFT (Pa- rameter Efficient Fine Tuning), complemented by LoRA (Low-Rank Adaptation) techniques. These advancements culminate in the creation of two dis- tinctive models: the proprietary GPT-4 and the open-source LLama-2.

### 3.1 Long Short-Term Memory Networks(LSTM)

Long Short-Term Memory (LSTM) networks are an enhanced form of recurrent neural networks (RNNs), specifically developed to tackle the prob- lem of learning long-range dependencies in se- quence data. LSTMs are highly effective in se- quence prediction tasks due to their unique internal structure, which includes multiple gates that man- age the flow of information. This design helps LSTMs to preserve information over prolonged pe- riods, avoiding the typical data loss seen in standard RNNs.(Chauhan, 2019)

The main advantage of Long Short-Term Mem- ory (LSTM) networks over conventional Recurrent Neural Networks (RNNs) is their ability to over- come the vanishing gradient problem. In traditional RNNs, the gradient often decreases sharply as it moves backward through the sequence, making it hard for the network to learn and remember past in- puts. LSTMs counter this problem with a gated ar- chitecture that maintains a stable gradient, thereby improving the network's ability to retain and learn from earlier information.(Miguel, 2021)

Architecture The architecture of an LSTM net- work is defined by several key components known as gates:

- **Forget Gate:** Determines which parts of the cell state should be discarded.
- **Input Gate (Learn Gate):** Decides what new information should be added to the cell state.
- **Cell State:** Acts as the internal memory of the LSTM, carrying information across the sequence processing.
- **Output Gate (Use Gate):** Regulates how much of the cell state is utilized to generate the output activation

2836

of the LSTM unit.(Nguyen, 2023)

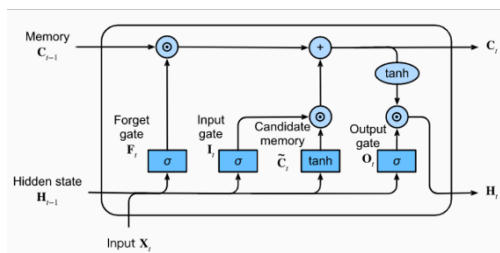

Figure 1: Illustration of the LSTM architecture (Al- mustafa, 2021).

### 3.2 RAG (Retreival Augmented Generation)

Retrieval Augmented Generation (RAG) is a frame- work that combines the strengths of retrieval sys- tems and generative language models. It aims to enhance the performance of language models by providing relevant contextual information from ex- ternal knowledge sources during the generation process. It consists of two main components: a re- triever and a generator. The retriever is responsible for retrieving relevant passages or documents from a knowledge base, given the input context. Vari- ous retrieval techniques can be used, such as sparse vector space models (e.g., BM25), dense vector rep- resentations (e.g., embeddings), or a combination of both. In our study we used embeddings. The retrieved passages are then fed into the genera- tor, which is typically a large pre-trained language model like GPT or BART. We used GPT in our study. The generator leverages the retrieved con- text to generate a more informed and knowledge- grounded output, drawing from the relevant in- formation present in the retrieved passages. The retrieved passages are concatenated with the in- put context and provided as input to the genera- tor. Compared to traditional language models that rely solely on their pre-trained knowledge, RAG models can potentially generate more accurate and informative responses by dynamically retriev- ing and incorporating relevant external knowledge. This is particularly beneficial in domains where fac- tual accuracy and knowledge grounding are crucial, such as question-answering, dialog systems, and knowledge-intensive applications. In our project, we use the RAG framework using GPT to enhance the performance of our conversational AI system for doctor-patient dialogues. By retrieving rele- vant medical knowledge from curated databases, the RAG model can provide more accurate and in- formative responses, drawing upon specialized do- main knowledge beyond what is captured in the pre- trained language model alone.(Lewis et al., 2021)

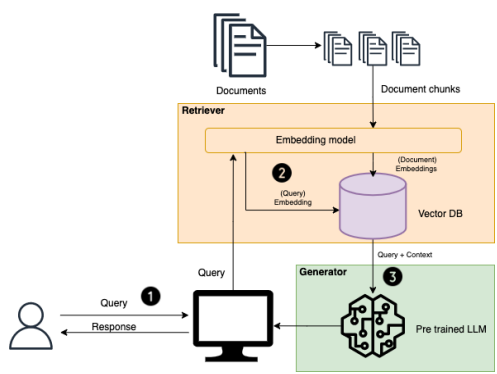

Figure 2: Illustration of the RAG architecure (Varkey, 2023).

### 3.3 Parameter-Efficient Fine-Tuning (PEFT) with Low-Rank Adaptation (LoRA)

As the Large Language models keep getting bigger (in terms of parameters), fine tuning them would incur high computational and memory costs as it necessitates updating all parameters which are bil- lions in numbers. To address this, the Low-Rank Adaptation (LoRA) technique is proposed for effi- cient fine-tuning of pre-trained language models.

2837

For a pre-trained weight matrix $W_0 \in R^{d \times d}$, its update is constrained by representing it with a low-rank decomposition $W_0 + \Delta W = W_0 + BA$, where $B \in R^{d \times r}$, $A \in R^{r \times d}$, and the rank $r \ll d$. During training, $W_0$ remains frozen and does not trainable parameters. Both $W_0$ and $\Delta W = BA$ receive gradient updates, while $A$ and $B$ contain are multiplied with the same input $x$, and their respective output vectors are summed coordinate- wise. The modified forward pass can be expressed as:

$$h = W_0 x + \Delta W x = W_0 x + BA x \qquad (1)$$

$A$ is initialized with random Gaussian values,

and $B$ is initialized with zeros, ensuring that

$\Delta W = BA$ is zero at the beginning of training.

$\Delta W x$ is scaled by $\alpha_r$, where $\alpha$ is a constant, and $r$

represents the rank.

The choice of "r" is a crucial for the LoRA al- gorithm to work because, tuning it very less would result is loss of crucial features because we would implictly remove linearly dependent features and by choosing a large "r", we increase the dimen- sion and increase the number of linearly dependent features.(Hu et al., 2021)

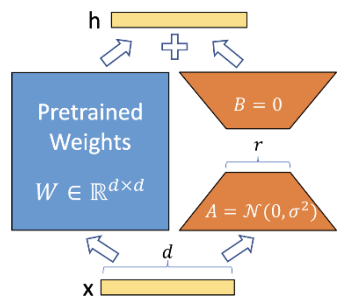

Figure 3: Illustration of the Low-Rank Adaptation (LoRA) technique (Hu et al., 2021).

### 3.4 GPT

In our project, we leverage the power of GPT for two primary purposes: Retrieval Augmented Gen- eration (RAG) and fine-tuning. Our approach in- volves utilizing a custom dataset comprising con- versations between patients and doctors to answer the questions for the given input.

#### 3.4.1 RAG using GPT

Prior to utilizing GPT-3.5 model, we employ the langchain framework's RecursiveCharacter- TextSplitter to chunk the preprocessed dataset. This text splitter is chosen for its ability to chunk the text while preserving semantic meaning within each chunk. This step is crucial for maintaining con- text and coherence during subsequent processing. Once the dataset is appropriately chunked, we uti- lize the OpenAI text embedding model to embed each chunk. These embeddings capture the seman- tic representation of the text and facilitate efficient comparison and retrieval. The embedded chunks are then stored in a vector database (vectorDB) for retrieval and further processing. We opt for ChromaDB as our vectorDB solution due to its lightweight nature and suitability for local storage. When a user submits a query, it undergoes the same embedding process using the OpenAI embedding model. The embedded query is then compared against the embedded chunks stored in the vec- torDB using cosine similarity. The top four most similar chunks are retrieved from the vector space. These retrieved chunks serve as contextual anchors for the subsequent response generation step. With the relevant chunks retrieved, we feed them into the GPT-4 large language model. Leveraging its contextual understanding and generation capabili- ties, GPT-4 generates a human-readable response based on the retrieved context. This

2838

response re- flects the model's comprehension of the user query within the broader conversation context, thereby enhancing the chatbot's conversational quality and relevance.

### 3.4.2 GPT Fine-tuning

For finetuning, the only option is to use OpenAI's fine-tuning API. As the GPT is a proprietary model, the weights and internal implementations like ar- chitecture, number of multi-heads are unknown. The dataset was prepared in a format suitable for the fine-tuning process, with each instance repre- sented as a dictionary containing "role" and "con- tent" keys. The "role" field distinguished between user queries (labeled as "user") and the desired assistant responses (labeled as "assistant").

Specifically, we fine-tuned the "gpt-3.5-turbo- 0125" model, a model of GPT optimized for fine- tuning tasks. The fine-tuning process was con- ducted for one epoch, with a batch size of 3 and a learning rate of 0.3. These hyperparameters were chosen to strike a balance between training effi- ciency and model performance.

After the fine-tuning process, we obtained a model ID representing the fine-tuned GPT-4 in- stance, tailored to our domain-specific dataset. This fine-tuned model was then loaded and utilized for generating responses to user queries, leveraging its acquired knowledge and capabilities specific to our conversational domain.(Brown et al., 2020)

### 3.5 Llama

In this project, we have leveraged the Open Source Llama-2 model, developed by Meta, to perform the task of conversation generation in the medical domain. We chose Llama-2 due to its extensive capabilities as an open-source model that focuses on stability, performance, and efficiency, making it well-suited for addressing diverse applications like text summarization, question answering, and dialogue generation. One of the key reasons for se- lecting Llama-2 over other Large Language Models is its introduction of Rotary Positional Embeddings, a novel approach to incorporating positional infor- mation into the model's self-attention mechanism. This technique has been shown to improve the model's performance, particularly in tasks involv- ing long-range dependencies and sequence model- ing.(Schick et al., 2020)

### 3.5.1 Rotary Positional Embeddings

Rotary Positional Embeddings (RoPE) is a tech- nique introduced in the Llama-2 model to encode positional information into the self-attention mech- anism. Unlike traditional positional encodings, which are added to the input embeddings, RoPE applies a rotary transformation to the query and key vectors in the self-attention layer.(Su et al., 2021)

Given a sequence of length $L$, RoPE generates two matrices $R_q \in R^{L \times d}$ and $R_k \in R^{L \times d}$, where $d$ is the dimensionality of the query and key vectors. These matrices are calculated as follows:

$$R^q(i,j) = \cos\left(\frac{i \cdot j}{10000^{\frac{j}{d}}}\right) \quad (2)$$

$$R^k(i,j) = \sin\left(\frac{i \cdot j}{10000^{\frac{j}{d}}}\right) \quad (3)$$

where $i$ represents the position index, and $j$ rep- resents the dimension index.

The query and key vectors, denoted as $q \in R^d$ and $k \in R^d$, respectively, are then transformed using the RoPE matrices:

$$q' = q \odot R_q(i, :) \quad (4)$$

2839

$$k' = k \odot R_k(i, :) \quad (5)$$

where $\odot$ represents element-wise multiplication,

and $i$ is the position index corresponding to the query and key vectors.

The transformed query $q'$ and key $k'$ vectors are then used in the self-attention computation, effec- tively encoding positional information into the attention mechanism. This approach has been shown to improve the model's performance in capturing long-range dependencies and handling tasks involv- ing sequence modeling, such as dialogue generation.

### 3.5.2 RoPE over Absolute and Relative Positional Embedding

Absolute Positional Embeddings (APE) assign a unique embedding to each token in a sequence based solely on its absolute position. While this approach proves effective for tasks where abso- lute order is critical (e.g., image captioning, lan- guage modeling), it presents limitations for tasks emphasizing relative positional relationships be- tween tokens (e.g., machine translation, question answering). APE struggles to capture these relative dependencies, potentially hindering performance in these domains.(Sinha et al., 2022)

Relative Positional Embeddings (RoPE) address this shortcoming by incorporating relative posi- tional information directly into the embedding space. Unlike APE, which assigns unique identi- fiers based on absolute position, RoPE captures the distance between tokens. This approach enables the model to grasp the sequential relationships within the input sequence more effectively. Studies have shown promising results with RoPE, particularly in tasks where relative position plays a significant role, leading to improved performance compared to APE.(Shaw et al., 2018)

Rotary Positional Embeddings (RoPE) offer a unifying approach by leveraging the strengths of both APE and RPE. This method mitigates the de- pendence on absolute position information solely, while simultaneously capturing the intricate posi- tional relationships between tokens. This balanced approach fosters the model's capacity to adapt to sequences of varying lengths and generalize effec- tively across diverse tasks. Consequently, RoPE emerges as a compelling choice for positional en- coding in sequence modeling applications.(Su et al., 2021)

### 3.5.3 Llama Fine-Tuning

For the Llama fine tuning we would be using all the above mentioned methods like PEFT with LoRa. So, we have choosen the model "meta-llama/Llama- 2-7b-chat-hf" which is an open source model com- prising 7 billion parameters. So, the perks of work- ing with the Open Source model, we can tune the architecture from scratch and perform very high level of customizations on the model. So, Initially we import the model using Auto Causal Model LLM.

For LoRa, we would be quantizing the dataset to a smaller bit configuration. Then we would tok- enize the dataset to the format Llama-2 accepts i.e. wrapping the prompt under the <sys> tag. Finally we defined the LoRa parameters (lora dropout, lora alpha) and also the training arguments. Then using the Supervised Fine Tuning Trainer (SFTTrainer) we train and fine tune the model according to our dataset, making the model proficient in the medical healthcare domain. (Yuan et al., 2023)

### 3.6 Llama RAG

Similar to GPT RAG, we also have implemented Retrieval Augmented Generation for the Llama-2 model (Meta). For this RAG aswell, we are uti- lizing the "meta-llama/Llama-2-7b-chat-hf" model provided by meta. Then after properly refining our dataset as mentioned in the Data Preprocess- ing stage of this paper, we collect this data into one knowledge base, we can think of it as a local folder directory for now. Understanding Context in Dialogue

2840

Systems. Effective natural language pro- cessing (NLP) tasks rely heavily on context to com- prehend and generate relevant responses. Large lan- guage models (LLMs) like Llama-2-7b have inher- ent limitations in context window size, restricting the amount of text they can process at once. In our system, we configure the context window to 4096 tokens, enabling the model to consider a significant textual scope when generating responses(Vaswani et al., 2017). However, for extended documents or conversational threads, splitting the input into smaller, manageable segments becomes necessary. This is where the LlamaIndex library plays a cru- cial role. It facilitates the chunking process and allows for efficient indexing and retrieval of rele- vant passages.

Sentence Transformers is a framework designed to compute dense vector representations of sen- tences and textual passages. Our system leverages the pre-trained all-mpnet-base-v2 model from Sen- tence Transformers to embed the textual data. Em- bedding refers to the process of transforming text into high-dimensional numerical vectors. These vectors hold significant value for various NLP tasks, including similarity search, clustering, and retrieval. The all-mpnet-base-v2 model's strength lies in its ability to generate high-quality sentence embeddings, capturing both semantic and syntactic information from the input text. By embed- ding our medical dialogue documents, we can efficiently index them and perform rapid similarity searches to retrieve relevant passages based on user queries.(Reimers and Gurevych, 2019)

The initial step in our system involves loading the medical dialogue documents from a designated directory using the Simple Directory Reader class. Subsequently, a LangchainEmbedding object is created. This object encapsulates the Sentence Transformers model, enabling it to generate em- beddings. The Llama Index library then utilizes these embeddings to construct a Vector Store Index. This data structure acts as a repository for stor- ing the documents and their corresponding embed- dings. The index facilitates efficient retrieval of relevant passages based on their degree of similarity to the user's query. We configure the index using a Service Context object, which specifies pa- rameters such as chunk size (set to 1024 tokens in our case), the chosen language model (Llama-2- 7b), and the embedding model (all-mpnet-base-v2). Finally, the index is persisted to a directory for fu- ture use, and a query engine is created. This query engine empowers users to interact with the index, enabling retrieval of relevant responses from the language model based on the retrieved passages.

### 3.7 Metrics

To evaluate the performance of our medical conver- sation chatbot, we used several widely-used metrics for assessing the quality of generated text. These metrics helps us to analyze by capturing different aspects of the generated responses, such as seman- tic similarity, fluency, and informativeness.

#### 3.7.1 BERT Score

The BERT Score is a metric that measures the se- mantic similarity between the generated text and the reference text by leveraging contextual embed- dings from the BERT model (Zhang et al., 2020). It computes a similarity score based on the cosine similarity between the token embeddings of the generated and reference texts, taking into account the contextual information. This metric provides a measure of the semantic coherence and relevance of the generated responses.

#### 3.7.2 BLEU Score

The BLEU (Bilingual Evaluation Understudy) score evaluates the quality of generated text by calculating the modified n-gram precision compared to the reference text. The BLEU score is computed as follows:

$$BLEU = BP \cdot \exp\left(\sum_{n=1}^{N} w_n \log p_n\right)$$

2841

where $N$ is the maximum length of n-grams con- sidered, $w_n$ are positive weights that sum to one, $p_n$ is the modified n-gram precision, and $BP$ is the brevity penalty that penalizes translations that are shorter than the reference text (Papineni et al., 2002). The BLEU score captures the fluency and adequacy of the generated responses by assessing the overlap of n-grams with the reference text.

### 3.7.3 ROUGE Score

ROUGE (Recall-Oriented Understudy for Gisting Evaluation) is a set of evaluation metrics that mea- sures the quality of a summary by counting the overlapping n-grams, word subsequences, or word pairs between the generated and reference sum- maries (Lin, 2004). While originally designed for evaluating summarization systems, ROUGE scores can also be applied to assess the informa- tiveness and relevance of generated text in other tasks, such as dialogue systems. Various ROUGE variants, like ROUGE-N (based on n-gram overlap) and ROUGE-L (based on longest common subse- quence), provide complementary insights into the quality of the generated responses.

## 4 EXPERIMENTATION

### 4.1 Experimentation Setup

To facilitate this experiment, we collected datasets from cited sources and utilized pre-trained Llama model weights from the Hugging Face repository. Our experimental setup used a range of tools and resources, including Jupyter Notebook and Kaggle for interactive coding and data exploration, Hiper- Gator for high-performance computing, and var- ious code libraries such as NumPy, TensorFlow, Pandas, LangChain, NLTK, Hugging Face API, CUDA, and ChromaDB. This comprehensive setup enabled us to test and evaluate the performance of both RAG and fine-tuning approaches in devel- oping an effective medical conversational chatbot. For the experimentation, we have used a novel and efficient database modeling for the efficient model retreival augmentation and response generation.

### 4.2 Datasets

For the project experimentation, we are using the combination of 3 datasets which are listed as below

**Medical-Dialog-Dataset**: https:// huggingface.co/datasets/medical_dialog

**Mohammed-Altaf's-Medical-Instruction-Dataset**: https://huggingface. co/datasets/Mohammed-Altaf/ medical-instruction-120k?row=13

**Diagnoise-Me-Dataset**: https://www.kaggle. com/datasets/dsxavier/diagnoise-me

### 4.3 Pre-Processing

To prepare the dataset , we initially preprocess the conversational data in four datasets into a uni- fied dataset in a structured format. Each conversa- tion snippet is encapsulated within tags denoting whether it's the patient's query or the doctor's re- sponse, resulting in a structured format like <Pa- tient> "Patient query"</Patient><Doctor>"Doctor Response"</Doctor>. This way included the mix of different types of doctor conversations for exam- ple, general physician, gynecologist, pediatrician etc. This efficient mix of the datasets provides the model with the knowledge of not just one domain, but a complete medical domain, to cover the maxi- mum amount of ground.

### 4.4 Experimentation Results

The average scores for various metrics are as shown in Table 1. These scores have been evaluated on different data points from the dataset and averaged the scores of each datapoint. Figure 4 explains the averaged scores

2842

visually for each model.

|  | BLEU | ROGUE | BERT | | |
|---|---|---|---|---|---|
|  |  |  | F1 | Precision | Recall |
| **LSTM** | 0.0037 | 0.031 | 0.209 | 0.177 | 0.258 |
| **GPT (FT)** | 0.372 | 0.294 | 0.584 | 0.616 | 0.571 |
| **GPT (RAG)** | 0.243 | 0.235 | 0.529 | 0.506 | 0.551 |
| **Llama (FT)** | 0.241 | 0.186 | 0.81 | 0.829 | 0.838 |
| **Llama (RAG)** | 0.288 | 0.259 | 0.861 | 0.851 | 0.875 |

Table 1: Metrics for different models

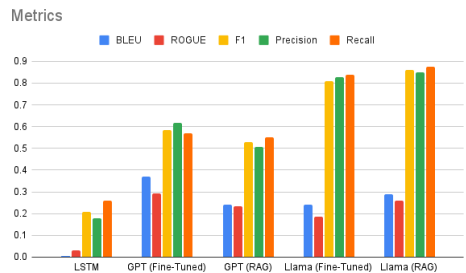

Figure 4: Metrics Bar Chart.

**4.5 Llama Training Process**

Llama being an Open Source model enabled us to go in-depth and assess its performance during train- ing, process(Figure 5, 6, 7). So we have recorded the whole training process of Llama, this not only helped us figure out best hyperparameters for the ef- ficient training process, keeping in check the GPU and compute power availability.

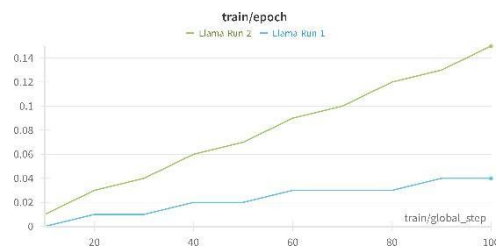

Figure 5: global step tradeoff with epochs

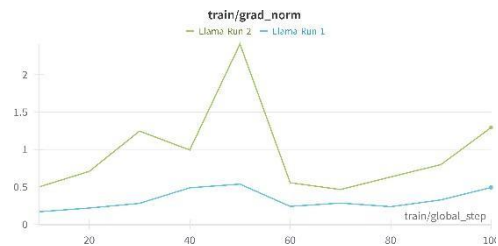

Figure 6: global step tradeoff with grad norm

2843

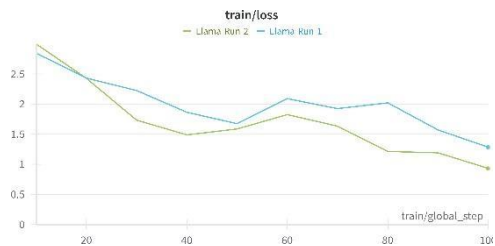

Figure 7: global step tradeoff with training loss

### 4.6 Experimentation Results

This novel approach of training the Large Language Model on datasets of multiple mixed expertise is expected to improve the model's robustness on new and unseen data. This method of training involved training the model on a large dataset, which made the LSTM model unable to properly approximate the training dataset, thus explaining the poor per- formance of LSTM models. Moreover, the training dataset contained very long prompts or the doc- tor/patient dataset, which made the LSTM unable to approximate and learn the long-term dependen- cies. The GPT model, although not an open-source model, due to the introduction of multi-head at- tention layers, which allowed it to preserve the long-term dependencies, explaining the effective performance of the GPT model compared to LSTM networks. Transitioning to the Llama model, its effectiveness can be attributed to several factors. Firstly, the introduction of rotary positional embed- dings (RoPE) addressed the challenge of encoding long-term dependencies in sequences, similar to the multihead attention mechanism in GPT. By in- corporating positional information directly into the self-attention mechanism, Llama was able to effec- tively capture the contextual relationships between tokens, enabling it to generate coherent and contex- tually relevant responses.

Moreover, the fine-tuning process of Llama on mixed-expertise datasets contributed to its robust- ness and adaptability to diverse conversational con- texts in the medical domain. Training the model on a large dataset comprising conversations between patient-doctors with varying levels of expertise en- sured that it could handle a wide range of queries and responses encountered in real-world scenarios. This approach not only enhanced the model's per- formance on new and unseen data but also enabled it to generalize better across different domains and conversation styles.

Furthermore, the retrieval-augmented generation (RAG) framework employed in Llama leveraged external knowledge sources to enhance the genera- tion process. By retrieving relevant passages from curated databases or previous conversations, Llama could enrich its responses with domain-specific information, improving the accuracy and informa- tiveness of the generated text. This dynamic inte- gration of external knowledge distinguished Llama from traditional language models and contributed to its effectiveness in medical dialogue systems.

### 5 CONCLUSIONS AND FUTURE WORK

In our evaluation of LSTM, GPT, and Llama mod- els, we observed distinct performance differences across various evaluation metrics. While LSTM exhibited poorer performance compared to large language models, GPT demonstrated superior co- herence and structured responses, outperforming Llama in metrics like BLUE and ROUGE scores. Conversely, Llama's strength lies in its ability to retrieve data with greater precision and generate

2844

responses more similar to the input, as indicated by higher BERT scores. This disparity can be attributed to several factors, including the intro- duction of rotary positional embeddings (RoPE) in Llama, enhancing its capacity to capture long- term dependencies similar to multihead attention layers in GPT. Additionally, fine-tuning Llama on mixed-expertise datasets contributes to its robust- ness and adaptability across diverse conversational contexts, addressing the limitations of traditional LSTM networks. Furthermore, Llama's retrieval- augmented generation framework enriches its re- sponses with domain-specific knowledge, distin- guishing it from conventional language models and enhancing its effectiveness in medical dialogue systems. Overall, our novel approach of training Llama on mixed-expertise datasets, combined with RoPE and retrieval-augmented generation, yields a robust and adaptable conversational AI model suit- able for various healthcare applications, overcom- ing the limitations of traditional LSTM networks and leveraging the strengths of advanced models like GPT.

The scope of this project can be extended across various dimensions due to its countless applications. One avenue is the development of an End-to-End mobile application leveraging powerful GPUs in the backend to provide a real-time chatting experi- ence with patients. This initiative would not only enhance the quality of life but also disseminate medical knowledge effectively.

The chat data gathered could serve as a basis for mood analysis of the patients, detecting signs of sadness or suicidal tendencies. Additional multi- heads could be employed to provide empathetic responses tailored to the patient's behavior, thereby boosting morale and aiding in the identification and management of chronic depression.

However, as the AI generates responses au- tonomously, there is a risk of transmitting inappro- priate information to unintended audiences, such as conveying highly technical data to users with limited domain knowledge or profane information to the wrong age groups. To mitigate this, robust and trusworthy AI algorithms and firewalls must be developed to ensure the information remains controlled and secured.